\documentclass[letterpaper, 10 pt, conference]{ieeeconf}  

\IEEEoverridecommandlockouts                              
\usepackage{graphics} 
\usepackage{epsfig} 
\usepackage{mathptmx} 
\usepackage{times} 
\usepackage{amsmath} 
\usepackage{amssymb}  
\usepackage{booktabs}
\usepackage{algorithm}
\usepackage{algpseudocode}
\usepackage{graphicx}

\usepackage{amsthm}
\usepackage{enumerate}
\usepackage{hyperref}
\usepackage{bm}
\usepackage{subcaption}
\usepackage[inline]{asymptote}
\usepackage[normalem]{ulem}
\usepackage{booktabs}
\usepackage{textcomp}
\usepackage{graphicx}

\title{\LARGE \bf
Transferable Task Execution from Pixels through Deep Planning Domain Learning
}

\author{Kei Kase$^{1,2}$, Chris Paxton$^{1}$, Hammad Mazhar$^{1}$, Tetsuya Ogata$^{2}$, Dieter Fox$^{1}$
\thanks{$^{1}$NVIDIA, USA
        {\tt\small {{kkase, cpaxton, hmazhar, dieterf}} at nvidia.com}}
\thanks{$^{2}$ Waseda University, Japan
        {\tt\small kase at idr.ias.sci.waseda.ac.jp, ogata at waseda.jp}}
}

{}

\begin{document}

\maketitle
\thispagestyle{empty}
\pagestyle{empty}

\begin{abstract}
While robots can learn models to solve many manipulation tasks from raw visual input, they cannot usually use these models to solve new problems. On the other hand, symbolic planning methods such as STRIPS have long been able to solve new problems given only a domain definition and a symbolic goal, but these approaches often struggle on the real world robotic tasks due to the challenges of grounding these symbols from sensor data in a partially-observable world. We propose Deep Planning Domain Learning (DPDL), an approach that combines the strengths of both methods to learn a hierarchical model. DPDL learns a high-level model which predicts values for a large set of logical predicates consisting of the current symbolic world state, and separately learns a low-level policy which translates symbolic operators into executable actions on the robot. This allows us to perform complex, multi-step tasks even when the robot has not been explicitly trained on them. We show our method on manipulation tasks in a photorealistic kitchen scenario.
\end{abstract}

\section{Introduction}
Much progress has been made on allowing robots to solve challenging problems purely from sensor data, e.g. RGB images and joint encoder readings~\cite{levine2016end,ghadirzadeh2017deep,duan2017one,huang2019neural,paxton2019prospection,Yu2018}. These approaches can be very reactive, and they operate on sensor data rather than needing models of the environment.
However, task specification remains a problem: even recent work in one-shot learning from demonstration focuses on tasks in visually simple environments, and requires a video demonstration of the task to be executed~\cite{duan2017one,xu2018neural,huang2019neural}.

Ideally, we would be able to reorder and recombine existing skills and employ them in new contexts to solve previously unseen problems with a different task structure, while still reacting to changes in the environment, all without ever having seen an instance of a task before -- something we refer to as transferable task execution. 

There exists one class of systems that in principle is very successful at solving this transferable task execution problem, given an appropriate problem domain $\mathcal{D}$: symbolic planners. These take problems specified in formal languages such as STRIPS~\cite{fikes1971strips} or PDDL~\cite{fox2003pddl2}, consisting of a set of operators $o$ with defined logical preconditions $L_P$ and effects $L_E$, and have been successfully applied to many problems, such as knitting, picking, and placing, in the past~\cite{balakirsky2012industrial,rovida2017extended,paxton2019representing}. Robust Logical-Dynamical Systems (RLDS), designed for reactive real-time execution of symbolic plans, add the run condition set $L_R$ and an associated control policy for each $o$. 
A variety of planning algorithms have been developed that can be applied to these problems, including FastDownward~\cite{helmert2006fast}.
In addition, the field of task and motion planning generalizes these problems to include continuous planning elements, e.g.~\cite{garrett2018stripstream,toussaint2018differentiable}.

\begin{figure}[tpb]
  \includegraphics[scale=0.57]{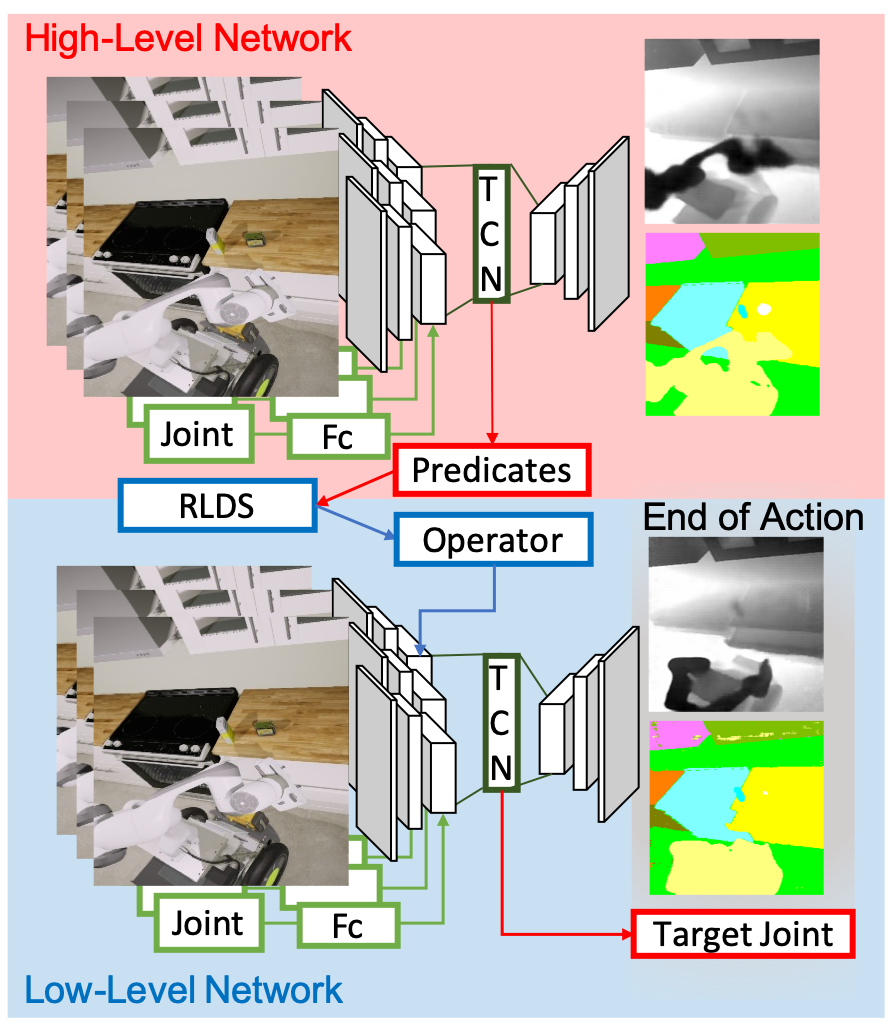}
  \caption{Overview of the framework for reactive task execution. The high-level network takes in sensor data and predicts logical world state $l$ consisting of a set of predicates $\rho$; these predicates are statements about the world like ``is the gripper around the sugar box'', associated with a truth value. The logical state is used with a symbolic planner to sequence actions for robust execution by a low-level policy.}
  \label{fig:System Overview}
  \vskip -0.5cm
\end{figure}

However, all of these systems suffer from a major shortcoming: defining the mapping between the planning domain and the real world, observed via sensors like cameras, is difficult. To this end, we propose Deep Planning Domain Learning (DPDL), a hierarchical approach which grounds a set of predicates $\rho$ making up the current logical (or symbolic) state of the world $l$ from images and other raw sensory data, and reactively chooses which operator -- learned sub-policy -- to execute by selecting the highest-priority enterable operator from a given task plan, as per~\cite{paxton2019representing}. An overview of this approach is shown in Fig.~\ref{fig:System Overview}.

Previous work has examined learning representations of complex, multi-step tasks~\cite{duan2017one,huang2019neural,huang2019continuous};
these approaches have been limited either by the need for a video demonstration including an action sequence~\cite{duan2017one,huang2019neural} or assume access to perfect state information and do not learn how to execute~\cite{huang2019continuous}.
By predicting the logical state, DPDL breaks up the problem into determining which of a subset of logical policies to execute, which lets us robustly execute an unseen task in a new environment.

To summarize, our contributions are:
(1) an approach for grounding symbolic predicates for use in task planning;
(2) a method for execution of unseen tasks given sensor data and a task description expressed as a goal set of logical predicates in a certain domain;
(3) and validation on a set of simulation tasks in a photorealistic simulation setting.

\section{Related Work}
Recently, there has been increasing interest in adding planning and symbolic reasoning into neural networks~\cite{srinivas2018universal,tamar2016value,garnelo2016towards,karkus2017qmdp}. Garnelo et al. learn a mapping between a learned symbolic state and actions for simple reinforcement learning tasks~\cite{garnelo2016towards}. Value iteration networks~\cite{tamar2016value} can learn to plan in 2D; QMDP-Net~\cite{karkus2017qmdp} adds the ability to plan under partial observability. Planning for a manipulation task in a complicated domain involves substantially larger state and action spaces, and adds considerable challenge to the problem. Srinivas et al. propose Universal Planning Networks~\cite{srinivas2018universal} focused solely on motion planning.

Some of the most relevant prior work has been done in one-shot learning from demonstration~\cite{duan2017one,huang2019neural,xu2018neural,Yu2018,huang2019continuous}.
Neural task programming~\cite{xu2018neural} proposes a reactive approach, modeling the problem using a detailed program trace.
Neural Task Graphs~\cite{huang2019neural} model the problem of one-shot imitation as one of graph generation. However, because the internal representation of the actions is not strongly supervised, it requires a full video demonstration of the task beforehand.
Yu et al.~\cite{Yu2018} proposed a method of training a set of primitives for multi-staged tasks; they combine primitives via a conditional high-level policy that uses a video demonstration from a human to specify the overall task.

More work has begun to leverage symbolic planning to help improve deep reinforcement learning~\cite{yang2018peorl,lyu2019sdrl}. Groshev et al.~\cite{groshev2018learning} proposed Generalized Reactive Policies, which are bootstrapped from symbolic planners and use them as a heuristic for solving more challenging planning problems. In our case, we are more interested in learning the problem of \emph{grounding} these problems, and rely on existing work for planning and execution -- though certainly these are complementary. These also look at simple, 2D, fully observable tasks~\cite{lyu2019sdrl,groshev2018learning}. In the future we could use our system's symbolic state predictions to train low-level RL policies as well.

Other work has looked at symbol grounding for robot planning and execution~\cite{abdo2012low,asai2017classical,dearden2014manipulation,paxton2019visual,paxton2019prospection,huang2019continuous}.
One interesting approach that does address this grounding problem is LatPlan, which performs unsupervised symbolic planning in a latent space~\cite{asai2017classical}, but explored only in fully observable 2D environments.
Huang et al.~\cite{huang2019continuous} proposed Symbol Grounding Networks, which work via a continuous relaxation of the planning problem, but do not explore partial observability or reactive execution.
Visual Robot Task Planning learns an unsupervised autoencoder representation for task planning~\cite{paxton2019visual} and other work incorporates natural language~\cite{paxton2019prospection} but these assume full observability; it's also difficult for these reconstruction-based models to learn a model of the logical effects of an action for planning.
To our knowledge, ours is the first work in this area to learn an explicit deep representation in photorealistic environments.

\section{Methods}
In this section, we present our proposed framework for hierarchical reactive control to accomplish tasks using just pixels and other raw sensor data. We look specifically at problems that can be expressed in a symbolic planning language, e.g. STRIPS~\cite{fikes1971strips} or PDDL~\cite{fox2003pddl2}.
A task planner such as FastDownward~\cite{helmert2006fast} can then compute an executable plan as a list of symbolic operators given a current logical state $l$ and a logical goal $L_G$; however, in our case, we assume we are operating in a partially-observable setting and do not have access to either the true underlying world state or the true logical state.

\subsection{Problem Definition}

Consider a Partially-Observable Markov Decision Process $(S, X, A, P, R)$, with states $s \in S$, actions $(a \in A)$, observations $x \in X$, transitions $P$ and reward function $R$. We are interested in solving arbitrary POMDPs with the same $(S, X, A, P)$ but with varying rewards given by $R$, where $R \in \{0, 1\}$ is determined by whether or not we have reached some logical goal $L_G \subseteq l$ consisting of a set of predicates describing properties of the environment.

Our representation of the world is based on the Robust Logical-Dynamical System~\cite{paxton2019representing}, which is itself based on common symbolic planning languages~\cite{fikes1971strips,fox2003pddl2}. We assume a planning domain $\mathcal{D}$, which contains a set of symbolic operators $o \in \mathcal{O}$. Operators might be, for example, ``Approach Obj.''
Each operator has associated logical preconditions $L_P$, run conditions $L_R$, and effects $L_E$, and defines a low-level policy $\pi^*(s) \rightarrow a$.

We can use the preconditions and effects, together with any known symbolic planner (such as FastDownward~\cite{helmert2006fast}) to create a \emph{robust logical-dynamical chain}, a sequence of states that we can reactively execute to achieve good performance even in the case of interference or perception noise~\cite{paxton2019representing}.
Each logical state is itself a set of predicates $\rho$, where each $\rho$ has different arguments consisting of symbols. A example predicates and operators are shown in Table~\ref{Tab:domain}.

Take $l_t = L(s_t)$ to be the logical state operator, which computes the logical state $l_t$ from the underlying world state $s_t$,
which we do not have access to at run time. Instead, we want to learn some function $f$ over our observation history $\vec{x}$ so that:
\[
    f(\vec{x}_{t}) \approx L(s_t)
\]
\noindent where $\vec{x}_t = \{x_{t-N}, \dots, x_t\}$ for some reasonable time window of length $N$. Given a logical state and an ordered list of operators $\vec{o}$, we choose the next operator whose $L_P$ (or $L_R$, if it is the current operator) are met.
Since we can compute the correct operator directly from $L(s_t)$ given $\vec{o}$, we will call the combination of $f$ and $\vec{o}$ our \emph{high-level policy} $\pi_{hl}$.

In addition, each logical operator $o_i \in \mathcal{O}$ represents a subset of the total action space $A$, so that many different continuous actions $a \in A$ correspond to a particular operator. Each $o_i$ is associated with some perfect ground-truth policy $\pi_i^*(s_t)$, which given a particular $s_t$ will compute the correct action for that state. In the same vein, however, we do not have access to $s_t$ at execution time. Therefore, we want to learn an approximate policy over $\vec{x}$:
\[
    \pi(\vec{x}_t | o_i) \approx \pi_i^*(s_t) = a^*_t
\]
\noindent where $a^*_t$ is the optimal action. Here, $\pi$ is a a deep neural network that will compute the correct continuous action to approximate the expert $\pi_i^*$, given knowledge of what the correct operator should be. We refer to this as the \emph{low-level policy} $\pi_{ll}$.

\subsection{High-Level V-TCN}

Our high-level policy $\pi_{(hl}$ utilizes a Variational Temporal Convolutional Network (V-TCN) that predicts the task-relevant predicates from a sequence of observations $\vec{x}_t$. 
The observations $x$ consist of RGB images from a camera mounted in the environment and joint-state readings from the robot, so $x_t = (I_{rgb,\ t}, q_t)$.

All our predicates are supposed to be a function of the true underlying world state $s_t$, so we model the latent world state in a similar way to a variational autoencoder~\cite{kingma2014auto}, as has been applied to robotics in previous work~\cite{ghadirzadeh2017deep}. We train a function that maps from observations to a Gaussian distribution over latent world states $\hat{s}_t$:
\begin{align*}
    f_{enc}(\vec{x}_t) &\rightarrow \mathcal{N}(\mu, \sigma)
\end{align*}
\noindent where $\mu_t, \sigma_t$ are parameters of the distribution and $\hat{s}_t \sim \mathcal{N}(\mu, \sigma)$ is sampled via the reparameterization trick. We add a KL divergence loss between $\mathcal{N}(\mu_t, \sigma_t)$ and $\mathcal{N}(0, I)$ to regularize the learned distribution.

\begin{figure}[btp]
  \vskip 0.1cm
  \includegraphics[scale=0.40]{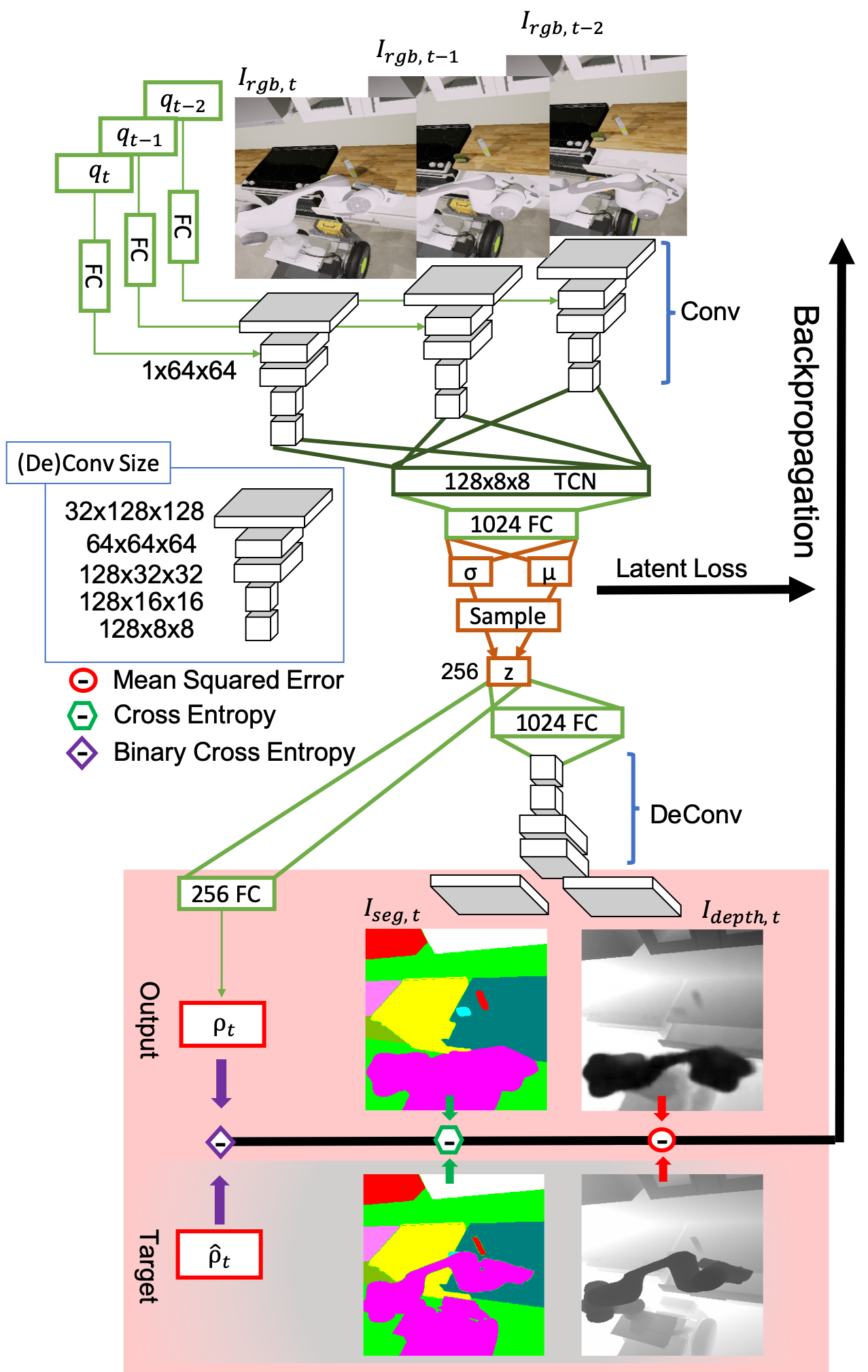}
  \caption{Overview of the High-Level V-TCN. This model predicts the current logical world state $L$, composed of a set of predicates $\rho$. A particular operator $o$ is selected based on this current logical state, by choosing the highest-priority $o$ whose preconditions $L_P$ or run conditions $L_R$ are met. It is then executed via the low-level V-TCN.}
  \label{fig:high-level}
  \vskip -0.5cm
\end{figure}

We then train a decoder $f_{dec}(\hat{s}_t)$ to predict the current logical state $l_t$.
The network also produces two auxiliary outputs: depth image $I_{depth,\ t}$, and semantic segmentation image $I_{seg,\ t}$. 
For our experiments, we chose a high-level time window of size $N=3$: at each time step $t$, the network intakes three of the most recent RGB image data $I_{rgb,\ (t-2):t}$ and robot arm joint angles $q_{(t-2):t}$.
The High-Level V-TCN is shown in Fig.~\ref{fig:high-level}.

When training the high-level model, we provide an auxiliary loss on estimating $I_{depth}$ and $I_{seg}$ so that our network will learn to capture spatial relationships between entities and identify objects.
We trained on the L2 loss to the ground-truth depth image $\hat{I}_{depth,\ t}$ and the cross-entropy loss to the ground-truth image $\hat{I}_{seg,\ t}$.

The final loss function $c$ is given by:
\begin{align*}
    c(\vec{x}_t) = &\lambda_{depth} \|I_{depth,\ t} - \hat{I}_{depth, t}\|_2 \\
    &+ \lambda_{seg} CE(I_{seg,\ t}, \hat{I}_{seg,\ t}) \\
    &+ \lambda_\rho BCE(\rho_t, \hat{\rho_t}) \\
    &+ 
    \lambda_{kl} D_{kl}\left(\mathcal{N}(\mu_t, \sigma_t), \mathcal{N}(0, I)\right)
\end{align*}
\noindent where $\lambda_{depth}$, $\lambda_{seg}$, $\lambda_\rho$, and $\lambda_{kl}$ are hyperparameters, $CE$ is the cross-entropy loss, and $BCE$ is binary cross entropy. In our experiments the $\lambda_{depth}$, $\lambda_{seg}$, $\lambda_\rho$, $\lambda_{kl}$ were set to 1, $10^{-1}$, 10, and $10^{-6}$ respectively to adjust the weight ratios between each losses. 

Each of the temporal RGB images is encoded using convolutional layers. We use CoordConv for the first two convolutional layers to add additional information about spatial relationships~\cite{Liu2018}.
We use a single fully connected layer to increase the dimensionality of arm joint angles to match the encoded RGB image features. The encoded values of arm joint angles and RGB images are then concatenated and further encoded with more convolutional layers. As spatial information is critical, no pooling is used in this network. The encoded data from each time slice is then convoluted in the temporal dimension to capture temporal information.
The temporally encoded features are further encoded with dense layers before features are sampled to extract latent features similarly to Variational Autoencoder. 

The latent features are used to compute boolean-valued predicates using dense layers. The latent features are also decoded using deconvolutional layers into current depth images and semantic segmentation images to improve training.

\subsection{Low-Level V-TCN}

\begin{figure}[btp]
  \vskip 0.1cm
  \includegraphics[scale=0.40]{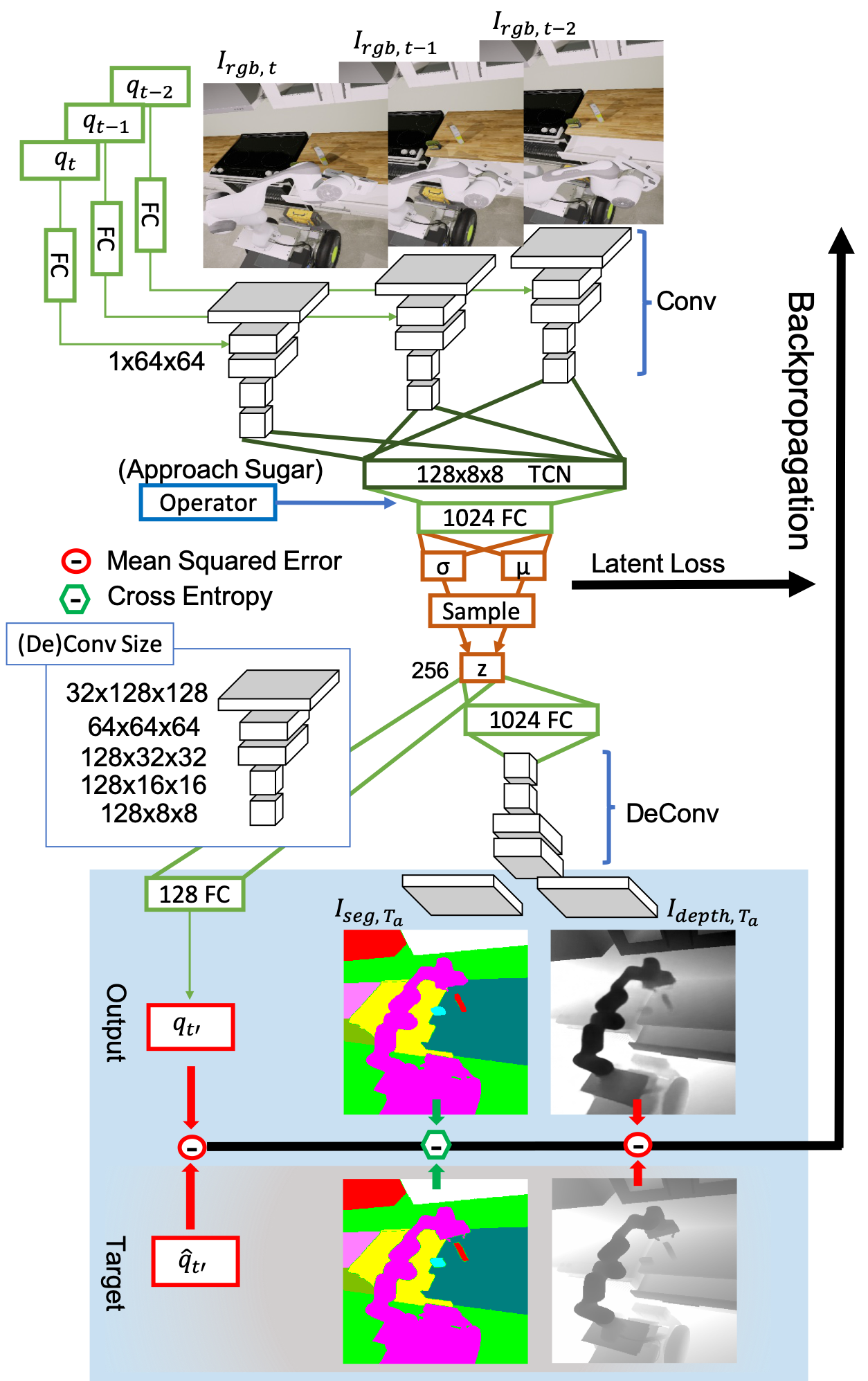}
  \caption{Overview of the Low-Level V-TCN. This model is a conditional policy, which executes a known operator such as ``Approach the sugar box.'' It generates a target configuration for the arm joint angles $q$ as well as depth and segmentation masks which can be used to generate motion plans.}
  \label{fig:low-level}
  \vskip -0.5cm
\end{figure}

Our low-level policy $\pi_{ll}$ is also modeled as a V-TCN. $\pi_{ll}$ predicts the action associated with the current operator $o$, where $o = \pi_{hl}(\vec{x})$. The network takes in multiple observations $x_t = (I_{rgb,\ t}, q_t)$ , in addition to $o$. Similar to our high-level model, it outputs $a$, $I_{depth}$, and $I_{seg}$ at time $t'$, the end of the action. 
In our case $a = q_{t'}$, the position in the robot's joint configuration space after the completion of $o$, but this does not need to be the case.
We use a Mean Squared Error Loss between $a$, given ground-truth actions $\hat{q}_{t'}$ from the dataset. The overview of the Low-Level V-TCN is shown in Fig.~\ref{fig:low-level}. 

\subsection{Execution}

Given these two models, we can perform execution of a new task given a particular environment. We take in sensor measurements and use them to determine our current logical state $l$, and use that to determine what operator we should execute.
Alg.~\ref{alg:execution} shows how this computation is performed and how our models are used in practice. 
A task planner computes an ordered list of operators to execute $\vec{o}$, from lowest to highest priority. At each step, the system computes all relevant predicates, then chooses the highest-priority operator (i.e., the one that is closest to the goal state, according to the task planner). This means that it will fluidly adapt if circumstances change.

There is one final consideration when reactively executing the plan $\vec{o}$. Since each operator can be sequenced in any order, we cannot simply execute any operator whose preconditions are met. Instead, once our symbolic planner has returned $\vec{o}$, we back-propagate preconditions from the goal state $L_G$ through the plan to enforce an ordering. Conditions are propagated backwards until they were created by an operator's effects; for more information, see prior work~\cite{paxton2019representing}. These extra conditions are simply added to each operator's $L_P$ and $L_R$.

\begin{algorithm}[bthp]
\caption{Algorithm to perform zero-shot task execution in a new environment, given domain $\mathcal{D}$ and  goal $L_G$}\label{alg:execution}
\begin{algorithmic}
\State \textbf{Given:} domain $\mathcal{D}$, 
\State $\vec{o} $ = \Call{Plan}{$\mathcal{D}$, $L_G$}
\While{not $L_G \subseteq l_t$}
\State $t = t + 1$
\State $x_t \leftarrow W$ // get latest world observation
\State $l_t \leftarrow \pi_{hl}(\vec{x}_t)$ // estimate current logical state
\State // follow prior work~\cite{paxton2019representing} to compute current operator
\For{$i \in (\text{length}(\vec{o}), \dots, 1)$}
    \If{$o_i \not= o_{t-1}$ and $L_P^i \subseteq l_i$}
        \State $o_t = o_i$; \textbf{break}
    \ElsIf{$o_{t-1} = o_{i}$ and $L_R^i \subseteq l_i$}
        \State $o_t = o_{t-1}$; \textbf{break}
    \EndIf
\EndFor
\State $q_t = \pi_{ll}(\vec{x}_t | o_t)$
\State \Call{GoTo}{$q_t$} // Execute motion 
\EndWhile
\end{algorithmic}
\end{algorithm}

\section{Experiments}
\begin{figure*}[bt]
  \centering
  \vskip 0.2cm
  \includegraphics[width=2.01\columnwidth]{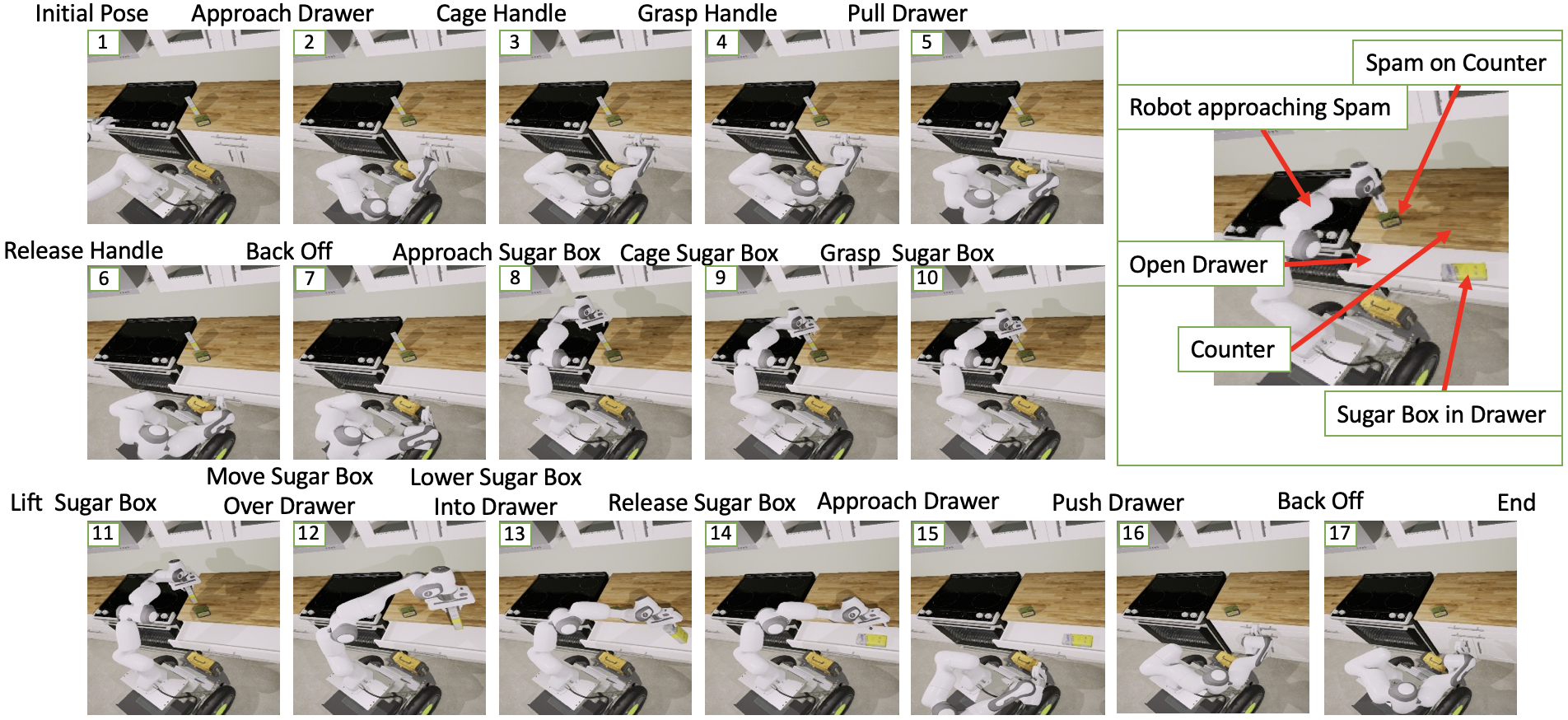}
  \caption{Overview of the Kitchen task. The array of scenes show the sequence of actions robots need to take for the put away sugar task. Each scene image is the view at the end of action labeled at left top and start of the action labeled at right top. The robot can put either object in the drawer, and must be able to do so from a relatively challenging viewpoint so that it can see the entire scene.}
  \label{fig:Kitchen Explanation}
  \vskip -0.3cm
\end{figure*}

To evaluate our framework, we perform a robotic manipulation task in a simulated kitchen domain environment (Fig.~\ref{fig:Kitchen Explanation}). The goal of the task is to put objects on the kitchen counter into the drawer. For the robot to complete the task, the robot must open the drawer, pick up the object, place it in the drawer, and close the drawer. We used a 7-DoF Franka Emika Panda for the experiments. 

We focus on two objects from the YCB dataset~\cite{ycb}, both placed randomly on the counter: a spam can and a sugar box. The system must determine specifically which stage of the task it should execute: if it begins to approach an object wrong, for example, it should back off and try again; if it drops an object accidentally, it should pick it up. 
An overview of the predicates comprising our logical state and the available operators is given in Table~\ref{Tab:domain}. A total of 42 predicates and 21 operators were used. 

\begin{table}[t]
\caption{List of predicates $\rho$ and operators $o$ in the Kitchen domain. The current logical state $l_t$ is composed of all these $\rho$ evaluated for all possible objects; $o$ can also act on multiple objects.}
\label{Tab:domain}
\begin{tabular}{l l}
\toprule
Predicates & Actions \\ \hline
Arm Is Above Counter & Open Gripper \\
Arm Is Around Handle Loose & Approach Drawer \\
Arm is Around (Obj or Handle) & Cage Handle \\
Arm Is Free & Grasp Handle \\
Arm Is Moving & Pull Drawer \\
Arm Is Near Handle & Release Handle \\
Arm Is Attached to Obj & Back Off \\
Arm In Approach (Obj or Handle) Region & Approach Obj \\
Arm In Driving Posture & Cage Obj \\
Arm In Front of Drawer & Grasp Obj \\
(Arm or Handle or Obj) Is Attached & Lift Obj \\
(Arm or Obj) Is Clear Above Counter & Move Obj Over Drawer \\
(Arm or Obj) Is In Drawer & Lower Obj Into Drawer \\
Drawer Is Open and Detached & Release Obj \\
Drawer Is Open & Approach Drawer \\
Drawer Is Closed & Push Drawer \\
Gripper Is Open &  \\
Obj Is On Counter &  \\
Obj Is Over Drawer &  \\
Obj Is Detected &  \\
Obj Is Tracked &  \\
\bottomrule
\end{tabular}
\vskip -0.5cm
\end{table}

The training dataset is collected in Programming by Demonstration fashion: the robot executes the task based on a policy which was manually engineered and logger logs the execution of the task. The logger collects RGB image, depth images, semantically segmented images, robot joint angles, current action, current predicates, and poses of the objects at 5 Hz. The camera captures $I_{rgb}$ as 480x640 images, that were resized to 256x256 for the network. Both the High-Level and Low-Level V-TCN take the $N=3$ most recent examples. The joint measurement $q$ includes all 7 joint angles, plus two dimensions for the gripper.

The training dataset for this experiment includes the ``put away spam can'' task and ``put away sugar box'' task, but no cases where two objects were put away simultaneously. The dataset is augmented with examples of the ``open drawer'' and ``pick up'' sub-tasks on their own, since accuracy is very important. 
At the start of every trial during data collection, objects were randomly placed within an area on the counter or in the drawer, the drawer state and robot joint angles were randomized, and the robot base was randomly placed near the cabinet.

\section{Results}
We evaluated the algorithm's performance entirely in a photorealistic simulation environment. Execution is stochastic, as it is in the real world, which increases the importance of reactive execution and adaptation to the environment.

\begin{table*}[bt]
\vskip 0.3cm
\caption{Performance comparison of different methods on tasks observed in training data.}
\label{Tab:Results}
\centering
\footnotesize
\begin{tabular}{l l l l l l l l}
\toprule
 & Oracle & Oracle (low level) & Oracle (high level) & Pose Estimation & PoseCNN~\cite{xiang2018posecnn} & V-TCN & DPDL \\
 \midrule
\multicolumn{1}{l}{Open Drawer} & 100.0 \% & 95.0 \% & 75.0 \% & 100.0 \% & N/A  & 0.0 \% & 70.0 \% \\ 
\multicolumn{1}{l}{Pick Spam} & 100.0 \% & 100.0 \%  &  100.0 \% & 0.0 \% & 40.0 \%& 50.0 \% & 90.0 \% \\ 
\multicolumn{1}{l}{Pick Sugar} & 100.0 \% & 100.0 \% & 100.0 \% & 0.0 \% &  40.0 \% & 45.0 \% & 95.0 \% \\ 
\multicolumn{1}{l}{Put Away Spam} & 100.0 \% & 100.0 \% & 95.0 \% & 0.0 \% & 25.0 \% & 0.0 \% 　& 90.0 \% \\
\multicolumn{1}{l}{Put Away Sugar} & 100.0 \% & 95.0 \%& 80.0 \% & 0.0 \% & 25.0 \% & 0.0 \%  & 85.0 \% \\ \bottomrule
\end{tabular}
\end{table*}

\textbf{Component Task Execution}
We tested several sub-tasks in 20 different random environments. Specifically, we looked at opening the drawer, picking the spam or the sugar box, or the complete task sequence including closing the drawer at the end. Success rates are listed in Table~\ref{Tab:Results} for different cases.
We show ``oracle'' performance using perfect ground truth information in order to demonstrate how well the model could work as an upper bound. In addition, we show two extra ``oracle'' results: ``oracle (low level)'' where the low-level execution has access to perfect simulation state, and ``oracle (high level)'' where we use the ground-truth predicate implementation to compute the action to execute.

In addition, we compare against three baselines: Pose Estimation,  PoseCNN~\cite{xiang2018posecnn}, and Single V-TCN. First, we show a realistic baseline system, in line with what would be used in the real world. We train a model to estimate the poses of all objects in the world from the same dataset and use these poses to compute the logical state in the same way as the oracle models.
Second, we compare against PoseCNN, a well-established pose estimation system, previously used in real-world robot experiments~\cite{paxton2019representing}.
Third, we compare against V-TCN, which is a Low-Level V-TCN without an action input.
Finally, we show DPDL model trained end-to-end to perform the specific tasks.

\begin{figure}[tpb]
  \includegraphics[scale=0.35]{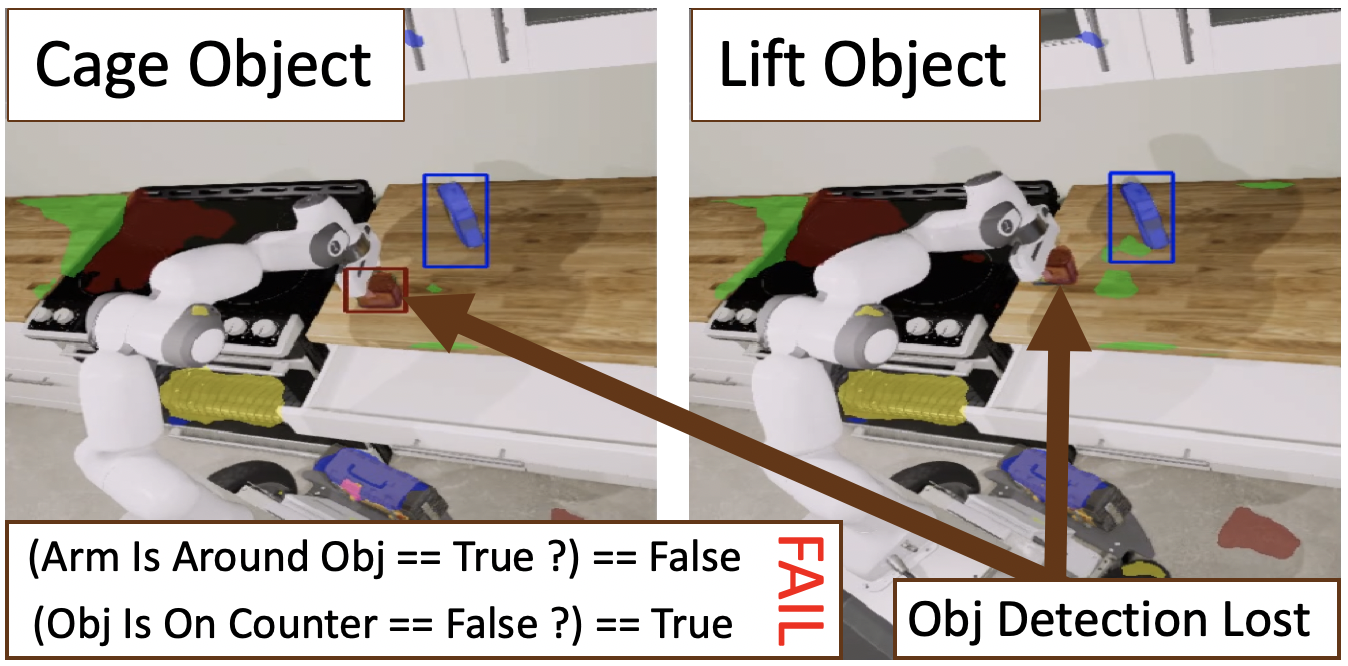}
  \centering
  \caption{Example of a case where PoseCNN~\cite{xiang2018posecnn} fails on our task. Even state of the art pose detection cannot usually handle occlusions, while end-to-end learning of predicate groundings allows us to still reactively execute a task plan.}
  \label{fig:PoseCNN_example}
  \vskip -0.5cm
\end{figure}

Results are shown in Table~\ref{Tab:Results}. With access to full ground-truth state information, the system performs perfectly; however, in a more realistic situation, we see the advantages of our approach. Using state-of-the-art perception such as PoseCNN~\cite{xiang2018posecnn} runs into issues because it's difficult to estimate the object's state when it is highly occluded (Fig. ~\ref{fig:PoseCNN_example}): failures occurred when either position estimates were slightly off. PoseCNN was trained on a synthetic dataset with much more variation for the sole purpose of estimating object poses, and its pose estimates are generally quite accurate for our task. PoseCNN is not trained to estimate the pose of the drawer handle; therefore, we used ground-truth state information for the handle, and an open drawer task is not tested. Pose Estimation could not estimate the poses of the object well enough for it to work with the system, but is able to estimate handle poses well. 

\textbf{Reactivity.}
One major advantage of our DPDL approach is reactivity, so we designed an approach to test this in different scenarios. While the robot is caging a particular object  during the pick up task, we teleport the object to a random area on the counter. The robot must adjust to the new object position and complete the task. While the robot never experienced this disturbance during training, it is still able to complete the target task due to our structure. We tested the robot perform reactive pick up spam task with a single V-TCN policy trained end-to-end and with and DPDL. The single V-TCN had a 30\% success rate and DPDL has an 85\%  success rate out of 20 trials. 

\textbf{Reactive Zero-Shot Execution.} 
The robot is further tested to complete the target task of put away sugar box and potted meat can one after another. This task is never presented to the robot in the training set. However, since the actions are chosen based on predicates, our framework has the capability of putting away the objects in series despite this not having been seen before in the test set. 

\begin{figure}[tpb]
  \includegraphics[scale=0.53]{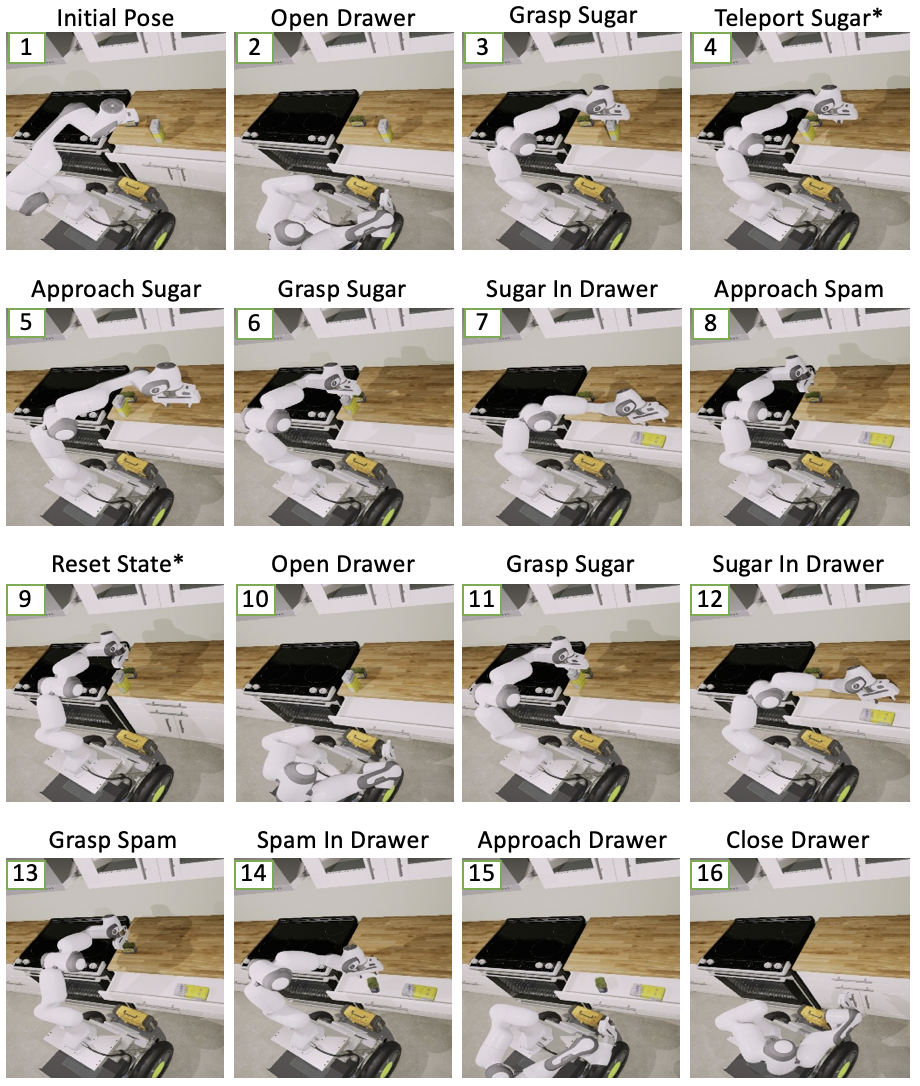}
  \caption{Reactivity test. The robot attempts to put sugar then spam in the drawer -- an unseen task. When it lifts the sugar, the sugar is teleported to a new location, and the robot must recover.}
  \label{fig:reactivity}
  \vskip -0.6cm
\end{figure}

Fig.~\ref{fig:reactivity} shows the full system, with a novel task (open drawer, move both objects, close drawer) and with the sugar box being teleported to a new location partway through the trial. Furthermore, the drawer is closed and sugar box is teleported back on the counter during the execution.

\section{Conclusions and Future Work}
We showed Deep Planning Domain Learning (DPDL), a system for zero-shot task execution that works by learning groundings for logical predicates from simulated training data. This system combines ideas from deep policy learning~\cite{huang2019neural} as well as classical planning~\cite{fikes1971strips,helmert2006fast,fox2003pddl2}.
In the future, we would like to use natural language to describe logical goals for actions, as in~\cite{paxton2019prospection}. We will also apply this work to real world scenes using domain randomization,
and greatly expand the problem domain. 

\bibliographystyle{IEEEtran}
\bibliography{main}

\end{document}